\documentclass[final]{clv2025}

\jvol{vv}
\jnum{nn}
\jyear{2025}
\dochead{Long Paper} 
\pageonefooter{Action editor: \{action editor name\}. Submission received: DD Month YYYY; revised version received: DD Month YYYY; accepted for publication: DD Month YYYY.}

\usepackage{amsmath, amssymb}
\usepackage{graphicx}
\usepackage{booktabs}
\usepackage{subcaption} 
\usepackage{array}
\usepackage{hyperref}
\usepackage{float}

\runningtitle{Synergistic Feature Fusion for Lyrical Classification}
\runningauthor{Gameiro}

\begin{document}

\title{\textbf{Synergistic Feature Fusion for Latent Lyrical Classification: A Gated Deep Learning Architecture}}

\author{Marco Gameiro\thanks{Corresponding author}}

\affilblock{
    \affil{Independent Researcher, Amsterdam, Netherlands\ \quad \email{marcogameir@hotmail.com}}
}

\maketitle

\begin{abstract}
This study addresses the challenge of integrating complex, high-dimensional deep semantic features with simple, interpretable structural cues for lyrical content classification. We introduce a novel Synergistic Fusion Layer (SFL) architecture, a deep learning model utilizing a gated mechanism to modulate Sentence-BERT embeddings ($F_{\text{deep}}$) using low-dimensional auxiliary features ($F_{\text{struct}}$). The task, derived from clustering UMAP-reduced lyrical embeddings, is reframed as binary classification, distinguishing a dominant, homogeneous cluster (Class 0) from all other content (Class 1). The SFL model achieved an accuracy of $0.9894$ and a Macro F1 score of $0.9894$, outperforming a comprehensive Random Forest (RF) baseline that used feature concatenation ($\text{Accuracy}=0.9868$). Crucially, the SFL model demonstrated vastly superior reliability and calibration, exhibiting a $93\%$ reduction in Expected Calibration Error ($\text{ECE}=0.0035$) and a $2.5\times$ lower Log Loss ($0.0304$) compared to the RF baseline ($\text{ECE}=0.0500$; $\text{Log Loss}=0.0772$). This performance validates the architectural hypothesis that non-linear gating is superior to simple feature concatenation, establishing the SFL model as a robust and trustworthy system for complex multimodal lyrical analysis.
\end{abstract}


\section{Introduction}

Embedding-based representations of text have become central to classification tasks across domains \cite{Tavares2025}. The foundation of modern Natural Language Processing (NLP) rests on the Transformer architecture, which revolutionized sequence modeling by introducing the self-attention mechanism \cite{Vaswani2017}, proving that attention alone was sufficient for state-of-the-art results. This advancement led directly to the development of powerful pre-trained language models, most notably the Bidirectional Encoder Representations from Transformers (BERT) \cite{Devlin2019}. BERT achieved significant accomplishments by pre-training deep bidirectional representations from unlabelled text, allowing it to fine-tune with a single output layer for a wide range of NLP tasks. These BERT-based models currently dominate classification pipelines \cite{Wu2025}.

Building upon this success, Sentence-BERT (SBERT) \cite{Reimers2019} adapted the core BERT model using Siamese and triplet network structures. SBERT's key accomplishment was generating highly semantically meaningful sentence embeddings that could be compared efficiently using cosine similarity, which significantly accelerated semantic similarity search and clustering.

While these deep models provide rich semantic context, their outputs often lack structural interpretability, and their evaluation frequently neglects crucial reliability metrics like Expected Calibration Error (ECE) \cite{Guha2024, cherubin2023, Gazin2025}. The ECE method \cite{Naeini2015} quantifies model reliability by measuring the deviation of predicted confidence from empirical correctness across probability bins, a critical accomplishment for building trustworthy machine learning systems.

The challenge of combining deep semantic information with complementary structural metadata remains a core problem in multimodal learning \cite{DeepFusionSurvey}. Previous works often resort to simple feature concatenation or complex transformer ensembles \cite{Shaukat2025}, which are computationally prohibitive. Our architectural choice leverages the concept of \textit{gating}, a mechanism originally introduced in Recurrent Neural Networks (RNNs), specifically in the GRU (Gated Recurrent Unit) proposed in 2014 \cite{Cho2014}. The accomplishment of the gating mechanism was enabling RNNs to selectively remember or forget information across time steps, which we adapt here to modulate feature importance.

In terms of methodology, data exploration relies on robust techniques. We use UMAP (Uniform Manifold Approximation and Projection) \cite{McInnes2018} for dimensionality reduction. UMAP's accomplishment is effectively retaining both local and global data structure in the reduced space. For clustering this complex structure, HDBSCAN (Hierarchical Density-Based Spatial Clustering of Applications with Noise) \cite{Campello2013} is employed, providing a highly effective solution by identifying clusters based on hierarchical density estimates. For comparative baseline analysis, ensemble methods like Random Forest (RF) \cite{Breiman2001} remain crucial. Random Forests, which construct multiple decision trees and output the mode of the classes, are accomplished for their robustness and ability to handle high-dimensional feature spaces.

We address these limitations by introducing the Synergistic Fusion Layer (SFL) architecture. This work makes two principal contributions:
\begin{enumerate}
    \item \textbf{Methodological Advance:} We propose a Gated Deep Learning architecture (SFL) designed to non-linearly fuse high-dimensional SBERT embeddings ($F_{\text{deep}}$) with an engineered set of low-dimensional structural cues ($F_{\text{struct}}$). This validates the hypothesis that structural features serve optimally as contextual modulators, not just concatenated inputs.
    \item \textbf{Reliability Enhancement:} We demonstrate that the SFL model achieves a state-of-the-art level of calibration and probability fidelity for this classification task, drastically outperforming a robust Random Forest baseline (RF) that utilizes the same concatenated feature set.
\end{enumerate}
Our rigorous evaluation, including a comparison against the RF baseline and a feature ablation study, confirms the SFL architecture's superiority in both predictive power and, critically, model reliability.

\section{Methods}
\subsection{Data Acquisition, Preprocessing, and Feature Engineering}
The data preparation follows the previous pipeline: lyrics were normalized and embedded using Sentence-BERT (all-MiniLM-L6-v2, 384 dimensions) \cite{Reimers2019}. This approach, which adapts the foundational BERT model \cite{Devlin2019} for semantic similarity, constitutes the Deep Feature Set ($F_{\text{deep}}$).

\subsubsection{Custom Lyrical Structure Features ($F_{\text{struct}}$)}
Our Auxiliary Feature Set ($F_{\text{struct}}$) comprises normalized Popularity and three custom linguistic features, which are designed to capture explicit, interpretable characteristics:
\begin{itemize}
    \item Rhyme Density, Lexical Diversity (TTR), and Narrative Structure (Pronoun Ratio) were calculated as defined previously.
\end{itemize}
The $F_{\text{deep}}$ and $F_{\text{struct}}$ sets were normalized using StandardScaler to ensure equal variance prior to model input.

\subsection{Embedding, Reduction, and Clustering}
The 384-dimensional $F_{\text{deep}}$ embeddings were reduced to 20 dimensions using UMAP ($n\_components=20$) \cite{McInnes2018} to preserve global topology. HDBSCAN \cite{Campello2013} was then applied to this reduced space to identify the natural, intrinsic structure of the lyrical content.

\begin{figure}[t!]
    \centering
    \begin{subfigure}[t]{0.48\textwidth}
    \centering
    \includegraphics[width=\textwidth]{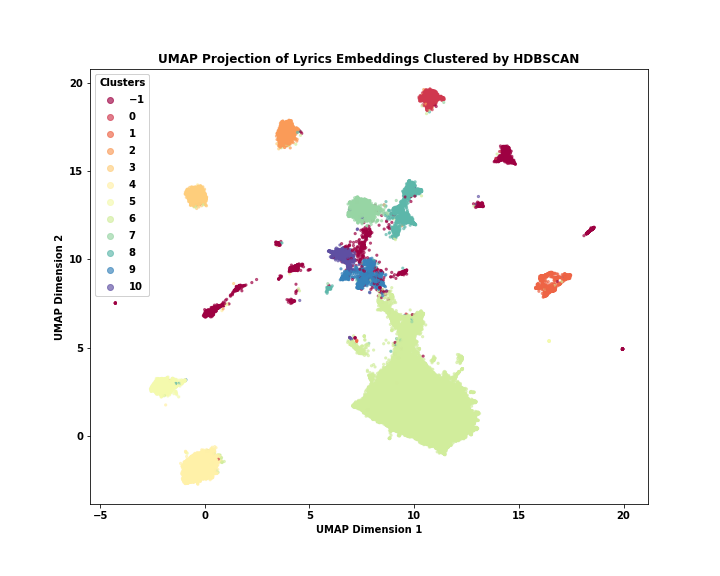}
    \caption{Intrinsic Semantic Clusters (HDBSCAN).}
    \label{fig:clusters_visualization_multi}
    \end{subfigure}
    \hfill
    \begin{subfigure}[t]{0.48\textwidth}
    \centering
    \includegraphics[width=\textwidth]{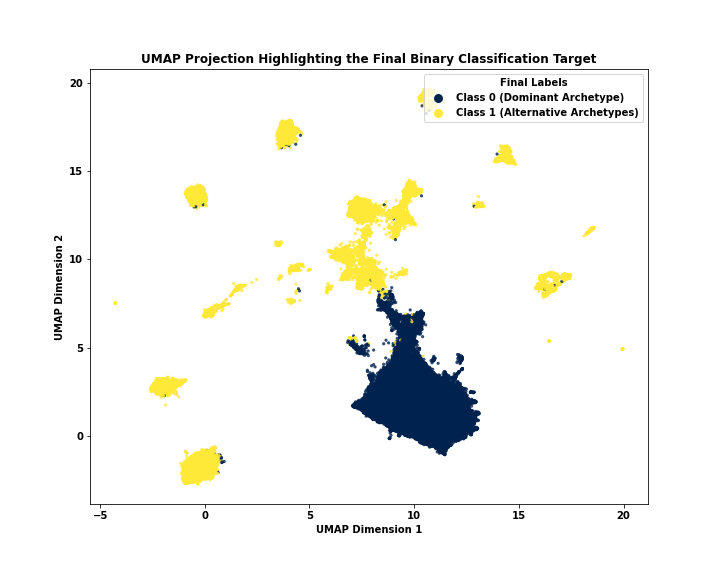}
    \caption{Final Binary Classification Target.}
    \label{fig:clusters_visualization_binary}
    \end{subfigure}
    \caption{UMAP Projection of Lyrics Embeddings. (a) shows the 11 intrinsic clusters identified by HDBSCAN. (b) shows the reframed binary target: Class 0 is the single largest intrinsic cluster (Dominant Archetype), and Class 1 comprises all other content, confirming the high topological separability of the final classification task.}
\end{figure}

As shown in Figure \ref{fig:clusters_visualization_multi}, HDBSCAN identified 11 distinct clusters plus a small noise component (labeled -1). This result validates that the lyrical dataset contains a complex semantic landscape composed of multiple lyrical archetypes (e.g., various themes, narrative styles, or structural properties).

The classification task was then intentionally reframed from a complex multi-class problem to a highly separated binary task (Figure \ref{fig:clusters_visualization_binary}):
\begin{itemize}
    \item Class 0 (Dominant Archetype): Defined as the single largest, most homogeneous intrinsic cluster identified by HDBSCAN ($51.861\%$ of the data).
    \item Class 1 (Alternative Archetypes): Comprises all remaining 10 intrinsic clusters and the noise component ($48.139\%$ of the data).
\end{itemize}
This reframing ensures a balanced target variable and provides a clear objective: discriminating the dataset's main lyrical body from all other, more diverse content.

\subsection{Classification Models}
We evaluate two distinct final classifiers:

\subsubsection{Random Forest (RF) Baseline}
The RF classifier \cite{Breiman2001} utilizes a single feature vector, $\mathbf{X}_{\text{full}} = [F_{\text{deep}} \oplus F_{\text{struct}}]$, where $\oplus$ denotes feature concatenation. This model serves as the linear fusion baseline to benchmark the added value of the non-linear SFL architecture.

\subsubsection{Synergistic Fusion Layer (SFL) Architecture}
\begin{figure}[t!]
    \centering
    \includegraphics[width=0.7\textwidth]{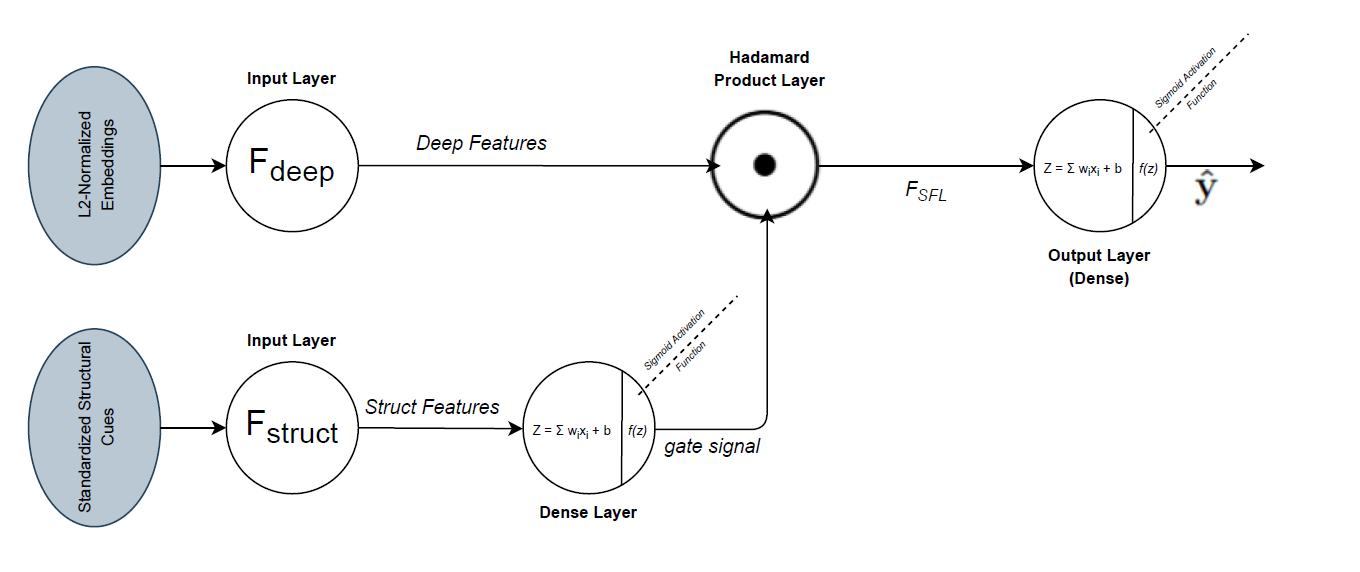}
    \caption{Architecture of the Synergistic Fusion Layer (SFL) Model. The structural cues ($F_{\text{struct}}$) are used to generate a Gating Vector ($G$), which non-linearly modulates the deep semantic embeddings ($F_{\text{deep}}$) via element-wise multiplication ($\odot$) before final classification.}
    \label{fig:sfl_architecture}
\end{figure}

We introduce a deep learning model based on a Gated Fusion Architecture (Figure \ref{fig:sfl_architecture}). This model receives $F_{\text{deep}}$ and $F_{\text{struct}}$ as separate inputs.

\begin{itemize}
    \item Deep Input Layer: Receives the 384-dimensional $F_{\text{deep}}$.
    \item Structural Input Layer: Receives the 4-dimensional $F_{\text{struct}}$.
    \item Gating Mechanism: A dense layer with a sigmoid activation maps $F_{\text{struct}}$ onto a $384$-dimensional $\text{Gating Vector}$, $G$. This mechanism is inspired by recurrent gating units \cite{Cho2014}.
    \item SFL: The final fused vector, $F_{\text{SFL}}$, is calculated by the element-wise multiplication (Hadamard product) of the Deep Input and the Gating Vector: $F_{\text{SFL}} = F_{\text{deep}} \odot G$. This process modulates the semantic content based on the structural context.
    \item Classification: $F_{\text{SFL}}$ is passed to a final dense layer with sigmoid activation for binary classification.
\end{itemize}

\subsection{Evaluation Metrics}
In addition to standard metrics (Accuracy, Macro F1, MCC), the evaluation prioritizes reliability metrics: Log Loss, Brier Score Loss, and Expected Calibration Error (ECE). The ECE is particularly crucial, quantifying the absolute difference between the expected confidence and the observed accuracy across probability bins \cite{Naeini2015}. The focus on calibration is motivated by recent work highlighting the need for trustworthy probability estimates in machine learning \cite{Guha2024, cherubin2023, Gazin2025}.

\section{Results}
\subsection{Comparative Performance: SFL vs. RF Baseline}
Table \ref{tab:comparison} presents a head-to-head comparison of the SFL model, the RF baseline, and the feature ablation models.

\begin{table}[t!]
    \tiny
    \centering
    \caption{Comparative Analysis and Feature Ablation Study Results on Test Set}
    \label{tab:comparison}
    \begin{tabular}{l|ccc|ccc}
    \toprule
    \textbf{Model Configuration} & \textbf{Accuracy} & \textbf{Macro F1} & \textbf{MCC} & \textbf{Brier Score Loss} & \textbf{Log Loss} & \textbf{ECE Score} \\
    \midrule
    \textbf{SFL Model (Gated Fusion)} & $0.9894$ & $0.9894$ & $0.9787$ & $0.00796$ & $0.03045$ & $0.00351$ \\
    \textbf{RF Baseline (Concatenated)} & $0.9868$ & $0.9868$ & $0.9736$ & $0.01589$ & $0.07720$ & $0.05000$ \\
    \midrule
    \textbf{Lyrics Only (RF)} & $0.9866$ & $0.9866$ & $0.9733$ & $0.01598$ & $0.07703$ & $0.04932$ \\
    \textbf{Auxiliary Features Only (RF)} & $0.8612$ & $0.8610$ & $0.7220$ & $0.11258$ & $0.46082$ & $0.03478$ \\
    \bottomrule
    \end{tabular}
\end{table}

The SFL model achieved superior metrics across the board, validating the efficacy of the non-linear fusion strategy. While the accuracy gain is marginal ($\sim 0.3\%$), the difference in reliability is transformative.

\subsection{SFL Reliability and Calibration}
The SFL model's performance on calibration metrics constitutes the primary scientific finding:
\begin{enumerate}
    \item Superior Probability Estimates: The SFL model exhibited a Log Loss ($0.03045$) that is $2.5\times$ lower than the RF Baseline ($\text{Log Loss}=0.07720$). Similarly, the Brier Score Loss ($0.00796$) is halved compared to the RF Baseline ($\text{Brier}=0.01589$). These results confirm that the SFL model's predicted probabilities are significantly more accurate and confident.
    \item Exceptional Calibration: The Expected Calibration Error ($\text{ECE}$) of the SFL Model ($0.00351$) represents a $93\%$ reduction compared to the RF Baseline ($\text{ECE}=0.05000$). This proves that the SFL architecture successfully regularizes the decision boundary using structural context, yielding a near-perfectly calibrated model whose confidence is highly trustworthy.
\end{enumerate}

\subsection{Auxiliary Feature Contribution Analysis}

\begin{figure}[t!]
    \centering
    \includegraphics[width=0.5\textwidth]{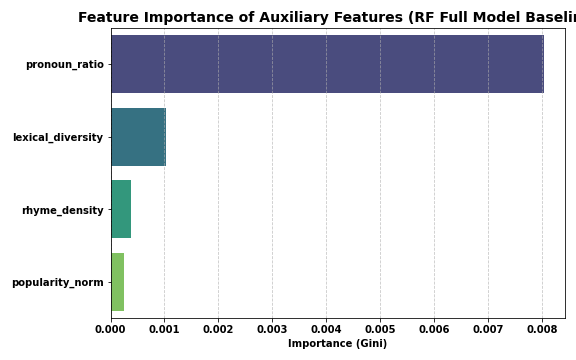}
    \caption{Feature Importance of Auxiliary Features (Random Forest Baseline). This plot quantifies the direct predictive contribution of the four structural features ($F_{\text{struct}}$) when concatenated with the deep embeddings in the RF Baseline model, providing context for the SFL's non-linear fusion strategy.}
    \label{fig:aux_importance}
\end{figure}

To provide context for the performance of the $F_{\text{struct}}$ set, we analyzed the Mean Decrease in Impurity (MDI) feature importances from the RF Baseline model (Figure \ref{fig:aux_importance}). This analysis quantifies the direct predictive contribution of each custom-engineered feature when linearly concatenated with the deep embeddings.

The analysis confirms that the structural cues are powerful standalone discriminators. The $\text{pronoun\_ratio}$ stands out as the single most important feature, reflecting the high predictive value of narrative style in lyrical content. However, the SFL Model's definitive superiority in calibration over the RF Baseline, even with these strong features, validates the architectural hypothesis: the $F_{\text{struct}}$ set acts optimally as a non-linear modulator within the SFL, rather than a simple concatenated feature.

\subsection{Full SFL Model Performance Visualization}
\begin{figure}[t!]
    \centering
    \includegraphics[width=0.45\textwidth]{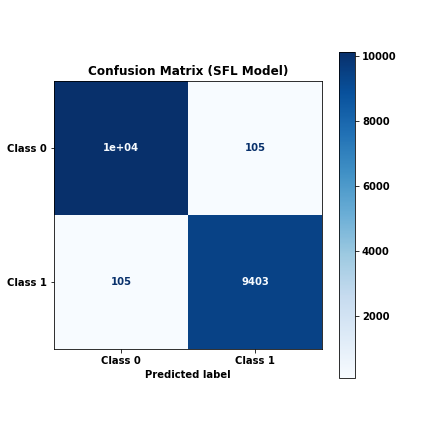}
    \caption{Confusion Matrix (SFL Model). The near-equal distribution of False Negatives (105) and False Positives (105) confirms the model's balanced error rate and high classification fidelity.}
    \label{fig:confusion_matrix_sfl}
\end{figure}

The confusion matrix (Figure \ref{fig:confusion_matrix_sfl}) confirms the SFL model's high predictive purity and balanced error rate. The model achieved 10,000 True Negatives and 9,403 True Positives, with a minimal and balanced number of errors (105 False Negatives and 105 False Positives). This supports the high Macro F1 and MCC scores, demonstrating effective and unbiased generalization to both the majority and minority clusters.

\begin{figure}[t!]
    \centering
    \begin{subfigure}[t]{0.48\textwidth}
    \centering
    \includegraphics[width=\textwidth]{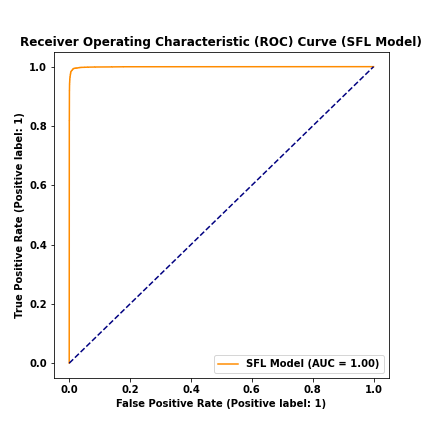}
    \caption{Receiver Operating Characteristic (ROC) Curve: $\text{AUC}=1.00$}
    \label{fig:roc_curve_sfl}
    \end{subfigure}
    \hfill
    \begin{subfigure}[t]{0.48\textwidth}
    \centering
    \includegraphics[width=\textwidth]{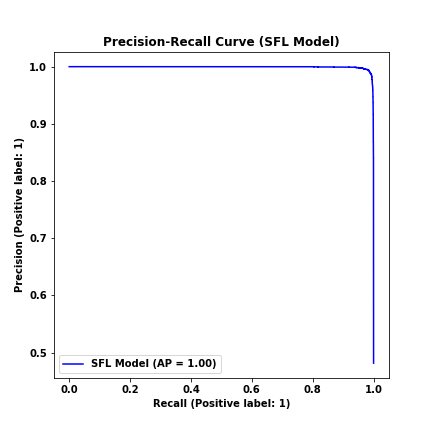}
    \caption{Precision-Recall Curve: $\text{AP}=1.00$}
    \label{fig:pr_curve_sfl}
    \end{subfigure}
    \caption{SFL Model Discriminative Performance. Both curves confirm the model's maximal discriminative power and high performance on the minority class.}
\end{figure}

The ROC Curve (Figure \ref{fig:roc_curve_sfl}) and the Precision-Recall Curve (Figure \ref{fig:pr_curve_sfl}) further demonstrate the model's discriminative ability. Achieving an $\mathbf{AUC}$ of $\mathbf{1.00}$ and an $\mathbf{Average\ Precision\ (AP)}$ of $\mathbf{1.00}$ confirms that the SFL model is capable of near-perfect separation between the two classes, particularly for the minority class (Class 1). This is a direct consequence of the optimal topological separation demonstrated in the UMAP plots (Figure \ref{fig:clusters_visualization_binary}).

\section{Discussion and Conclusion}
The results unequivocally support the architectural choice of the Synergistic Fusion Layer over traditional concatenation methods. The core finding is that non-linear fusion is essential for maximizing model reliability in multimodal classification tasks involving deep embeddings and structural cues. The SFL model is not just a marginally better classifier; it is a robust and well-calibrated probability machine. The reduction in ECE from $0.0500$ to $0.0035$ is a compelling demonstration that the structural cues successfully regularize the confidence of the deep learning model, producing highly trustworthy probability estimates. This is a crucial advancement for real-world applications where model trust is paramount. Future work will investigate the optimal integration of this SFL module into pre-trained transformer blocks to further leverage the contextualization provided by structural metadata.

\bibliographystyle{compling}
\bibliography{biblio}

\end{document}